\documentclass[11pt,a4paper]{article}
\usepackage[hyperref]{acl2020}

\usepackage{times}
\usepackage{latexsym}
\usepackage{graphicx}
\usepackage{bm}
\usepackage{verbatimbox}
\usepackage{booktabs}
\usepackage{amsmath}
\usepackage{multirow}
\usepackage{paralist}
\usepackage{enumitem}
\usepackage{xcolor}
\usepackage{amsfonts}

\DeclareMathOperator*{\softmax}{softmax}
\DeclareMathOperator*{\MLP}{MLP}
\DeclareMathOperator*{\Transformers}{Transformers}

\usepackage{microtype}

\aclfinalcopy 
\def\aclpaperid{511} 

\title{{P}rediction or {C}omparison: {T}oward {I}nterpretable {Q}ualitative {R}easoning}

\author{
Mucheng Ren,
    \ Heyan Huang$^{*}$\
    and
Yang Gao \\
School of Computer Science and Technology, Beijing Institute of Technology, Beijing, China \\
Beijing Engineering Research Center of High Volume Language Information \\ Processing and Cloud Computing Applications, Beijing, China \\
\texttt{\{renm,hhy63,gyang\}@bit.edu.cn} \\
}

\date{}

\begin{document}
\maketitle
\renewcommand{\thefootnote}{\fnsymbol{footnote}}
\footnotetext[1]{Corresponding author}
\renewcommand{\thefootnote}{\arabic{footnote}}
\begin{abstract}
Qualitative relationships illustrate how changing one property (e.g., moving velocity) affects another (e.g., kinetic energy) and constitutes a considerable portion of textual knowledge. Current approaches use either semantic parsers to transform natural language inputs into logical expressions or a ``black-box" model to solve them in one step. The former has a limited application range, while the latter lacks interpretability. In this work, we categorize qualitative reasoning tasks into two types: prediction and comparison. In particular, we adopt neural network modules trained in an end-to-end manner to simulate the two reasoning processes. Experiments on two qualitative reasoning question answering datasets, QuaRTz and QuaRel, show our methods' effectiveness and generalization capability, and the intermediate outputs provided by the modules make the reasoning process interpretable.
\end{abstract}

\section{Introduction}
\begin{figure}[t]
    \includegraphics[width= \linewidth]{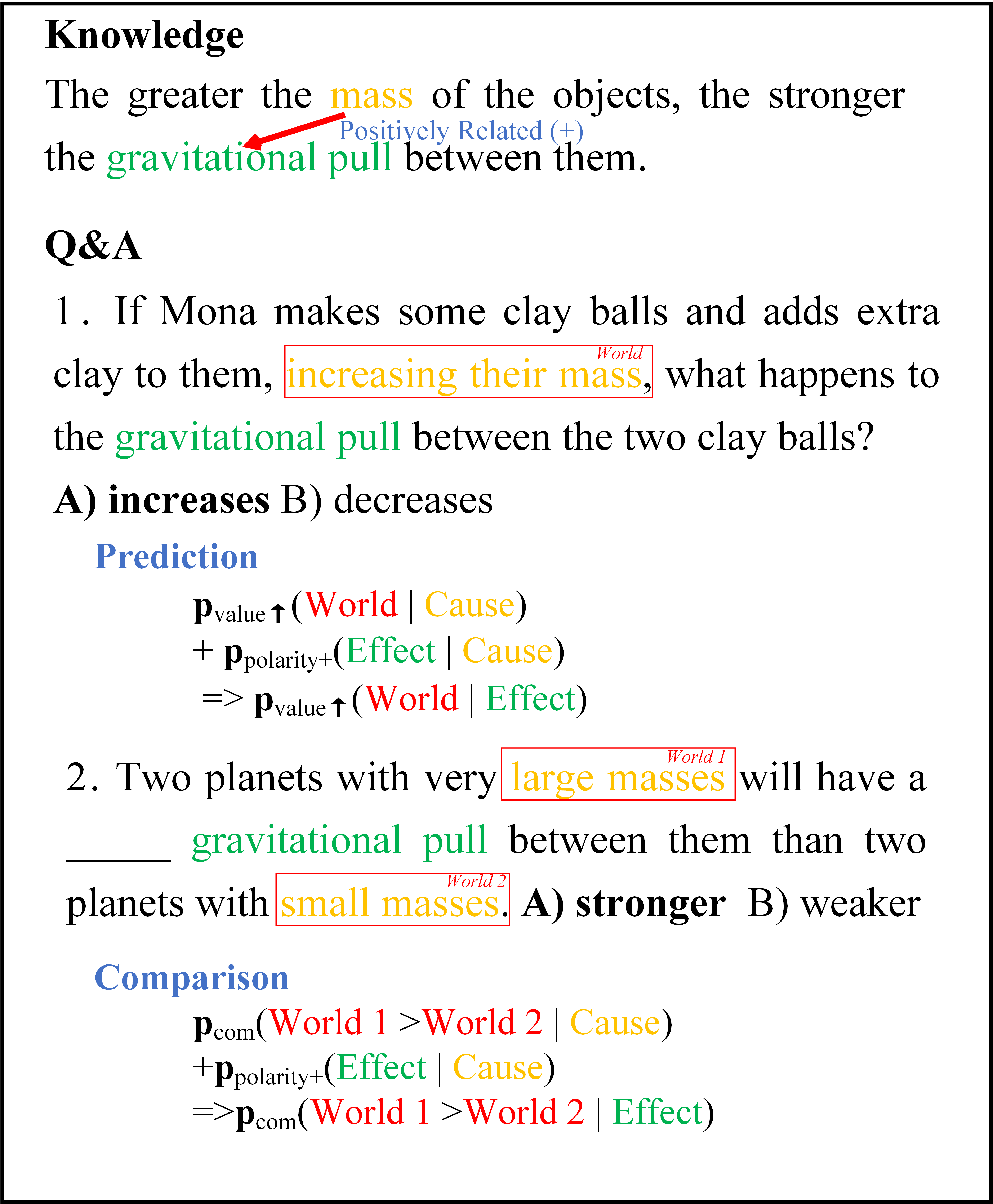}
    \caption{Examples from QuaRTz dataset.}
    \label{fig:example}
\end{figure}

Qualitative relationships abound in our world, especially in science, economics, and medicine. Since Question Answering (QA) has been significantly developed in recent years, various challenging datasets have been proposed~\cite{lai-etal-2017-race,rajpurkar-etal-2018-know,yang-etal-2018-hotpotqa,choi-etal-2018-quac,dua-etal-2019-drop}, and one often encounters the context and questions about qualitative relationships. Figure~\ref{fig:example} shows an example that requires reasoning about the qualitative relationship between the mass and the gravitational pull, where the knowledge sentence states that the mass is positively correlated with the gravitational pull, and the questions describe different scenarios that test the flexible application of the knowledge.
\par
Therefore, understanding the qualitative relationships behind the context and applying the qualitative textual knowledge for reasoning is essential. However, the most current datasets on qualitative reasoning are much smaller than standard Question Answering datasets, making the task challenging and receiving limited attention. At present, in the two mainstream qualitative relationship question datasets, QuaRel~\cite{tafjord2019quarel} and QuaRTz~\cite{tafjord-etal-2019-quartz}, the existing methods can be divided into two categories, symbolic reasoning based on a semantic parser~\cite{krishnamurthy2017neural,tafjord2019quarel} and a ``black-box" model based on representations~\cite{ Mitra2019AGA,tafjord-etal-2019-quartz,asai-hajishirzi-2020-logic,mitra-etal-2020-deeply}.  

\par
The two types of methods have their advantages and disadvantages. On one hand, symbolic reasoning provides solid explanations for the problem-solving process, but existing semantic parsers are trained to translate natural-language sentences into a task-specific logical representation that naturally increases the demand for additional annotated data and has limited generalization capability. On the other, approaches based on pre-trained language models that solve the task in one step achieve stunning results at the expense of limited interpretability.
\par
To tackle the issues mentioned above, in this paper, we propose a neural network module-based approach that provides good interpretability while achieving better results. Specifically, inspired by human cognitive processes, we group the qualitative reasoning questions into two categories: \textbf{Prediction} and \textbf{Comparison}. As the example in Figure~\ref{fig:example} shows, each category requires different reasoning chains. In \textbf{Prediction}, the question asks to directly predict the effect on an occurrence of the cause on an entity in the situation. In \textbf{Comparison}, the question asks to compare the effects on the two entities. Then we adopt neural modules to model each step in these two reasoning chains, and all modules are trained in an end-to-end manner.
\par
The practice of first summarizing the questions and then modeling the different reasoning chains step by step has three advantages: First, it has good generalization ability. The two reasoning chains summarized cover the vast majority of qualitative reasoning problems, so there is no need to design logical expressions for specific tasks. Second, it achieves better performance, because complex qualitative reasoning tasks are broken down into simple sub-tasks, with the modules required to complete only simple sub-tasks. Third, it provides better interpretability, because each module can provide a transparent intermediate output.
\par
Experimental results on the QuaRel~\cite{tafjord2019quarel} and QuaRTz~\cite{tafjord-etal-2019-quartz} datasets demonstrate the effectiveness and generalization capability of our proposed approach. It surpasses the state-of-the-art model by a large margin (absolute difference ranging from 1.8\% to 4.4\%). Furthermore, analyses of the intermediate outputs and a human evaluation show that each module in our approach performs well on its corresponding sub-task and clarifies interpretability.
\section{Related Work}
\paragraph{Qualitative Reasoning:} This type of reasoning is an indispensable part of artificial intelligence.~\citet{forbus1984qualitative} and~\citet{weld2013readings} propose formal models for qualitative reasoning when the research filed is emerging. However, there has been little work on reasoning with textual qualitative knowledge. \citet{tafjord2019quarel} solves such tasks using semantic parsing and a symbolic solver; this type of method naturally falls short on performance when compared to neural systems. Meanwhile, some works tackle tasks using a data-driven ``black-box" model based on representations computed by language models, which achieves superior performance~\cite{tafjord-etal-2019-quartz,Mitra2019AGA,asai-hajishirzi-2020-logic,mitra-etal-2020-deeply}. Our work enjoys the performance improvement that a neural network brings but also provides interpretability.
\paragraph{Neural Network Modules:} These modules were first adopted in the Visual Question Answering (VQA) domain ~\cite{johnson2017clevr,andreas2016neural,hu2017learning}. Due to the interpretable, modular, and inherently compositional nature of neural network modules, they were further extended to natural language~\cite{gupta-lewis-2018-neural,jiang-bansal-2019-self,jiang2019explore}.~\citet{gupta2019neural} extend neural module networks to answer compositional questions. ~\citet{ren-etal-2020-towards} and~\citet{liu2020multi} further introduce neural network modules on one complex reasoning task ROPES~\cite{lin-etal-2019-reasoning}. In this work, we explore the effectiveness of neural network modules on a qualitative reasoning task.
\begin{figure*}[ht]
    \centering
    \includegraphics[width=\textwidth]{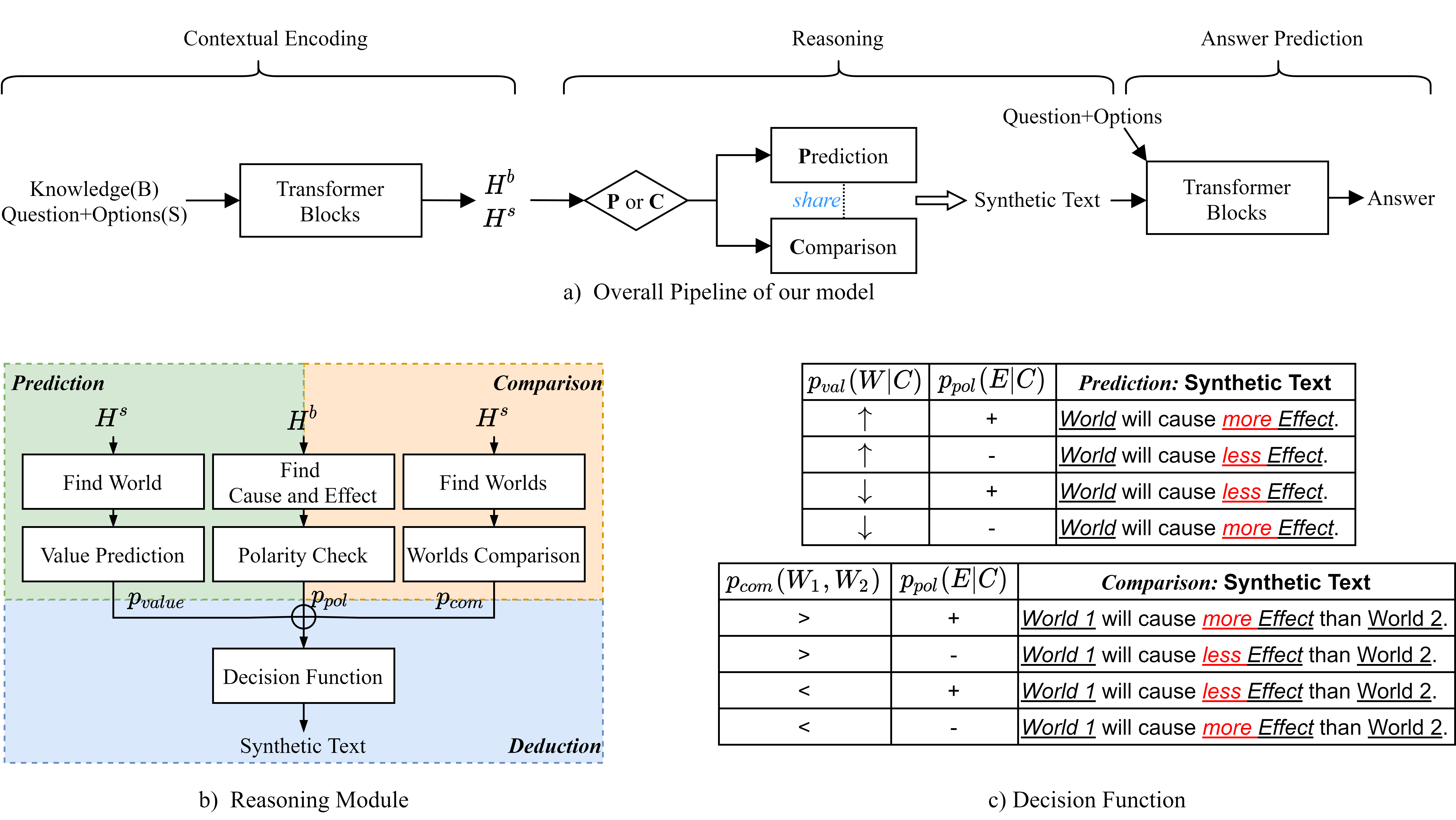}
    \caption{The overall diagram of our proposed model. Given background knowledge $B$ and statement $S$ formed by the question and options, we first generate the contextual representations. Then we synthesize a text by collecting the evidence provided by the \textbf{Prediction} or \textbf{Comparison} module, where the detailed structure is shown in b). Last, we predict the answer with the aid of the synthetic text.}
    \label{fig:overall}
\end{figure*}

\section{Method}
Our work solves the question through a three-step process illustrated in Figure~\ref{fig:overall}a:
\begin{enumerate}
    \item The \textit{Contextual Encoding} component encodes the concatenation of knowledge, the question, and options into contextual representations.
    \item The \textit{Reasoning} component arranges different reasoning chains according to the type of problem, where Prediction and Comparison contain multiple neural modules. A synthetic text is generated by the decision function based on the evidence collected from the reasoning chains.
    \item The \textit{Answer Prediction} component takes the concatenation of the question, options, and the generated synthetic text instead of given knowledge as the input and returns the answer option with the highest salience scores.
\end{enumerate}
\subsection{Contextual Encoding}
We selected RoBERTa~\citep{devlin-etal-2019-bert,liu2019roberta} as the encoder to encode the background knowledge, the question, and two answer options together and output contextualized embeddings. In particular, we concatenated the question with two answer options as statement $S =\left \{ s_{i} \right \}_{i=1}^{m}$, similar to the examples shown in Figure~\ref{fig:example}. Furthermore, we joined given background knowledge $B =\left \{ b_{j} \right \}_{j=1}^{n}$ and the statement with the special tokens used in language models as $\left \langle s \right \rangle b_1,\dots, b_n \left \langle \slash s \right \rangle \left \langle \slash s \right \rangle s_1,\dots,s_m \left \langle \slash s \right \rangle$, which was further fed into RoBERTa containing a series of successive transformer blocks,
\begin{equation}
    \bm{H^b,H^s}= \Transformers(B,S),
\end{equation}
where $\bm{H^b} \in \mathbb{R}^{n \times d}$, $\bm{H^s} \in \mathbb{R}^{m \times d}$ are contextual representations of the knowledge and the statement, respectively, and $d$ is the dimension for the hidden states. The representations $\bm{H^b,H^s}$ are padded into fixed length and served as the global variables in the \textit{Reasoning} part.

\subsection{Reasoning}
The architecture of the reasoning component is shown in Figure~\ref{fig:overall}b. In particular, we designed the different reasoning chains depending on the question type. Specifically, both reasoning chains were constructed by several end-to-end modules, and some modules were shared. Then the evidence collected from each module is summarized for the final deduction.
\subsubsection{Prediction }
\paragraph{Find Cause and Effect:}
To answer the Prediction-type question like the first example shown in Figure~\ref{fig:example}, we first need to understand what the knowledge describes, i.e., figure out the cause and effect properties (e.g., \textit{mass} and \textit{gravitational pull}). To achieve this, we applied a multilayer perceptron (MLP) over background knowledge representations $\bm{H^b}$ and then used a softmax function to normalize the projected logits and get attention scores over all knowledge tokens for the cause property and the effect property respectively,
\begin{align} 
 \bm{p}_{c}^{b}&=\softmax(\MLP(\bm{H^b};\theta_{c}))\in\mathbb{R}^{n}, \\
 \bm{p}_{e}^{b}&=\softmax(\MLP(\bm{H^b};\theta_{e}))\in\mathbb{R}^{n},
\end{align}
where $\bm{p}_{c}^{b}$ and $\bm{p}_{e}^{b}$ are the attention vectors over knowledge in terms of  the cause and effect properties, and $\theta$'s are trainable parameters in the MLP. Take Figure~\ref{fig:example} as an example: $\bm{p}_{c}^{b}$ is the attention over background tokens, whose value is much larger for \textit{mass} than the other tokens.

\paragraph{Polarity Check:}
We further need to judge the polarity of the qualitative relationship between the cause and effect properties (e.g., \textit{mass} and \textit{gravitational pull} are positively related). To achieve that, we first applied cause and effect property attention vectors on the knowledge representation $\bm{H^b}$ to obtain the weighted ones. Then we took the average of the weighted representations and concatenated them together. After that, another MLP followed by a softmax function computed the probabilities based on the representations,
\begin{gather}
    \bm{p}_{pol} =  \softmax(\MLP((\bm{{H^b}}^T\bm{p_{c}^{b}};\bm{{H^b}}^T\bm{p_{e}^{b}});\theta_{pol})),
\end{gather}
where $\bm{p}_{pol} = [\bm{p}_{pol+}$, $\bm{p}_{pol-}]$ denotes the probability of positive and negative correlations, $\theta_{pol}$ is a learnable parameter of the MLP. In the example shown in Figure~\ref{fig:example}, $\bm{p}_{pol+}$ is supposed to be larger than $\bm{p}_{pol-}$.

\paragraph{Find World:}
Furthermore, we should return to the statement and find out the world that happened in the statement (e.g., \textit{increasing their mass}). To achieve that, we used the same setting as  the\textit{Find Cause and Effect Property} module,
\begin{align} 
 \bm{p}_{w}^{s}&=\softmax(\MLP(\bm{H^s};\theta_{w}))\in\mathbb{R}^{m}, 
\end{align}
where $\bm{p_{w}^\bm{s}}$ is the attention vectors over the statement for the world, and $\theta$ is the learnable parameters of the MLP. 

\paragraph{Value Prediction:}
Moreover, we need to decide whether the attribute value in World is an increment ($\bm{p}_{val\uparrow}$) or decrement($\bm{p}_{val\downarrow}$; e.g., \textit{increase their mass} is an increment). To achieve that, we derive representations of the world by averaging statement representation $\bm{H^b}$ weighted by the world attention vector, $\bm{p_{w}^s}$. Then, another MLP was stacked with softmax to get the probabilities,
\begin{gather}
    \bm{p}_{val} =  \softmax(\MLP((\bm{{H^s}}^T\bm{p_{w}^{s}});\theta_{val})),
\end{gather}
where $\bm{p}_{val} = [\bm{p}_{val\uparrow}$, $\bm{p}_{val\downarrow}]$ denotes the probability that the attribute value is incremented or decremented, and $\theta_{val}$ is a learnable parameter of the MLP. In the example shown in Figure~\ref{fig:example}, $\bm{p}_{val\uparrow}$ is supposed to be larger than $\bm{p}_{val\downarrow}$.

\subsubsection{Comparison}
The Comparison reasoning chain contains two modules that are shared with the Prediction reasoning chain, \textit{Find Cause and Effect}, and \textit{Polarity Check}. However, the Comparison chain differs from the Prediction chain in two modules.
\paragraph{Find Worlds:}
In the Comparison-type question like the second example shown in Figure~\ref{fig:example}, the statement asks to compare the effects on the two entities. Thus, two worlds appeared in the statement instead of one (e.g., \textit{large masses} and \textit{small masses}). To achieve that, we took the same straightforward method as the \textit{Find World} module,
\begin{align} 
 \bm{p}_{w1}^{s}&=\softmax(\MLP(\bm{H^s};\theta_{w1}))\in\mathbb{R}^{m}, \\
  \bm{p}_{w2}^{s}&=\softmax(\MLP(\bm{H^s};\theta_{w2}))\in\mathbb{R}^{m}, 
\end{align}
where $\bm{p_{w1}^\bm{s}},\bm{p_{w2}^\bm{s}}$ are the attention vectors over the statement for worlds 1 and 2, and $\theta$'s are learnable parameters of the MLP. 
\paragraph{Worlds Comparison:}
This module aims to compare the worlds in terms of the cause property. For example, world 1 (\textit{larger masses}) is more relevant to \textit{the greater the mass} than world 2 (\textit{smaller masses}) in Figure \ref{fig:example}.
To achieve that, we adopted a bilinear function to evaluate the relevance between each world's cause property and the cause property in background knowledge, which is further normalized into a probability with softmax,
\begin{align}
    \bm{p}_{com} = \softmax(&(\bm{{H^b}}^T\bm{p}_{c}^{b})^{T}\bm{W}_{com}(\bm{{H^s}}^T\bm{p}_{w1}^{s}), \notag \\
    &(\bm{{H^b}}^T\bm{p}_{c}^{b})^{T}\bm{W}_{com}(\bm{{H^s}}^T\bm{p}_{w2}^{s})),
\end{align}
where $\bm{W_{com}}\in\mathbb{R}^{d\times d}$ is a trainable matrix, and $p_{com}$ denotes the probability that the world is relevant to the cause property. 
\subsubsection{Deduction}
The Deduction module aims to conduct the reasoning by considering the evidence computed from each module in the reasoning chain. In particular, we applied the decision functions as shown in Figure~\ref{fig:overall}c for each reasoning type to generate the synthetic text.
\par
We take the examples in Figure~\ref{fig:example} as an illustration. From the \textit{Find Cause and Effect}, and \textit{Polarity Check} modules, we conclude a positive relationship existed between the cause \textit{mass} and the effect \textit{gravitational pull}, so we denote $\bm{p}_{pol}(\bm{E}ffect|\bm{C}ause)$ as $+$. 

For Prediction, the \textit{Value Prediction} module states that the attribute value of the world in terms of cause property is an increment. Thus, we denote $\bm{p}_{val}(\bm{W}orld|\bm{C}ause)$ as $\uparrow$. Therefore, the world must also have an increment on the effect property due to the positive relationship between the cause and effect. In this way, we generated a synthetic text by slot-filling as follows: \textit{large mass} will cause \textbf{more} \textit{gravitational pull}.

Similarly, for Comparison, the \textit{Worlds Comparison} module showed that compared with world 2, world 1 was more relevant to the cause property in the background knowledge. Thus, we denote $\bm{p}_{com}(World 1, World 2)$ as $>$ (i.e., $\bm{p}_{com}(World 1|Cause) > \bm{p}_{com}(World 2|Cause))$. In this way, we rendered world 1 larger than world 2 in terms of the effect property because of the positive polarity. Then, a synthetic text was also be generated.

\subsubsection{Reasoning Type Classification}
\label{sec:classification}
As the reasoning type is not available during the inference time, we added a classifier based on the statement representations, which is defined as follows:
\begin{gather}
    \bm{s} = \frac{1}{m}\sum_{i}^{m}\bm{H_{i}^{s}}\in \mathbb{R}^{d},\\
    \bm{p}_{cla} = \sigma(MLP(\bm{s};\theta_{cla})),
\end{gather}
where $\bm{s}$ can be viewed as an embedding of the statement, $\theta_{cla}$ is the trainable parameter in the MLP, and $\sigma$ is an activation function. $\bm{p}_{cla}$ denotes the classification scores.
\subsection{Answer Prediction}
\label{sec:ap}
In the Answer Prediction component, we replaced the given background knowledge with the synthetic text from the Reasoning component. Then we combined it with the question and answer choice  $A = \{a_{k}\}_{k=1}^2$ as $\left \langle s \right \rangle Synthetic Text \left \langle \slash s \right \rangle\left \langle \slash s \right \rangle q;a_{k} \left \langle \slash s \right \rangle$. We used the final hidden vector corresponding to the first input token ($\left \langle s \right \rangle$) as the aggregate representation. We further predicted the probabilities of an answer being the answer choice $a_{k}$ in the same manner as in~\citet{liu2019roberta}.
\subsection{Model Training}
Two models (i.e., reasoning and answer prediction) are learned in our approach. 
\paragraph{Reasoning}
The loss function for reasoning is defined as
\begin{gather}
\label{eq:loss}
    \ell_{Reason} = -\sum_{\bm{y}\in \bm{Y}} \alpha_{\bm{y}} \gamma_{\bm{y}}\bm{\Tilde{y}}^{T}log(\bm{y}).
\end{gather}
$\bm{Y} = \{\bm{p_{w}^{s}}\in\mathbb{R}^{m},\bm{p_{w_1}^{s}}\in\mathbb{R}^{m},\bm{p_{w_2}^{s}}\in\mathbb{R}^{m},\bm{p_{e}^{b}}\in\mathbb{R}^{n},\bm{p_{c}^{b}}\in\mathbb{R}^{n}, \bm{p}_{val}\in\mathbb{R}^{2},\bm{p}_{com}\in\mathbb{R}^{2},\bm{p}_{cla}\in\mathbb{R}^{2} \} $ are predictions of different modules, $\bm{\Tilde{y}}^{T}\in \left \{ 0,1 \right \}^m$ or $ \bm{\Tilde{y}}^{T}\in\left \{0,1 \right \}^n$ or $ \bm{\Tilde{y}}^{T}\in\left \{0,1 \right \}^2$ are corresponding gold labels, $\gamma_{\bm{y}} \in\left \{0,1 \right \}^2$ denotes whether the loss function should use the label (as some annotations would be missed due to different reasoning type), and $\alpha_{\bm{y}}$ is the weight for each module $y$.

\paragraph{Answer Prediction:}
We took the standard binary cross entropy loss $\ell_{AP}$ as the training objective of the Answer Prediction component.

During the inference time, we first determined the reasoning type based on the classification score obtained in section~\ref{sec:classification}. Then we followed the Deduction module to synthesize corresponding text and further fed it into the trained Answer Prediction model. The answer choice with the highest probability was selected as the final answer. 

\section{Experiment Setup}
\subsection{Datasets}
Datasets QuaRTz~\cite{tafjord-etal-2019-quartz} and QuaRel~\cite{tafjord2019quarel}, were used to evaluate the proposed model. Both datasets require reasoning about textual qualitative relationships and provide well-defined annotations used for symbolic reasoning. The detailed statistics are in Table~\ref{tab:statistics}.
\paragraph{Auxiliary Supervision:}
The loss function for the Reasoning component takes two different types of supervision, i.e., span supervision (e.g., cause, effect spans in the knowledge) and binary logits supervision (value prediction, world comparison, and reasoning type). The QuaRTz dataset provides the annotations for the property descriptions and property values, but not every instance in the dataset contains such annotations (2280 out of 2696 are annotated completely). Furthermore, some annotations are not standard and must be further processed to satisfy the training objective. Therefore, we mitigated these issues by generating an additional auxiliary from hypothesis and manual annotating. The detailed guidelines and labeled examples are in Appendix~\ref{sec:ASC}. All modules in the Reasoning component are trained only on the QuaRTz (the proportion of Prediction and Comparison questions is 2296:400).  
\begin{table}[!t]
\centering
\begin{tabular}{@{}lccc@{}}
\toprule[2pt]
\textbf{Statistics }                & \textbf{Train} & \textbf{Dev} & \textbf{Test}  \\ \midrule
QuaRTz & 2,696  & 384 & 784 \\
QuaRel  & 1,941  & 278  &552 \\
\bottomrule[2pt]
\end{tabular}
\caption{Statistics of the datasets}
\label{tab:statistics}
\end{table}
\subsection{Implementation Details}
We used the pre-trained language model RoBERTa-large in the experimentation\footnote{\url{https://github.com/huggingface/transformers}}. In particular, we trained all modules in the Reasoning component end-to-end on one Nvidia RTX8000 48GB GPU and used two GPUs for the Answer Prediction component. We tuned the Reasoning component parameters according to the averaged performance of all modules and tuned the parameters in the Answer Prediction component based on the final accuracy. We selected F1 scores and accuracy as the evaluation metrics. Specifically, we set $\alpha_{\bm{y}} $'s in equation~\ref{eq:loss} as follows: 0.1 for span-based loss, 0.2 for the \textit{World Comparison, Value Prediction}, and \textit{Reasoning Type Classification}. The detailed hyperparameters and search bounds are described in the Appendix~\ref{sec:p_list}.

\subsection{Compared Models}
\paragraph{QUASP and QUASP+:} The models proposed by~\citet{tafjord2019quarel} which extend type-constrained semantic parsing to address the problem of
tracking different ``worlds" in questions.
\paragraph{BERT and RoBERTa:} The standard multiple-choice QA frameworks are based on a powerful pre-trained language model ~\cite{devlin-etal-2019-bert,liu2019roberta}. As described in section~\ref{sec:ap}, it took the concatenation of knowledge, the question, and answer options as the input and used the first token representation for classification.
\paragraph{gvQPS+:} Instead of introducing a semantic parser, this model applies the generate-validate framework~\cite{Mitra2019AGA}. It first generates a natural language description of the logical form and validates whether the natural language description is followed from the input text.
\paragraph{DeepEKR:} Similar to~\citet{Mitra2019AGA}, this model replaces the parser with a neural network, softening the symbolic representation~\cite{mitra-etal-2020-deeply}.
\paragraph{LG(DA,Reg):} The RoBERTa-based model leverages logical and linguistic knowledge to augment labeled training data (DA) and then uses a consistency-based regularizer (Reg) to help the training~\cite{asai-hajishirzi-2020-logic}.
\begin{table}[!t]
\resizebox{\linewidth}{!}{%
\begin{tabular}{@{}lcccc@{}}
\toprule[2pt]
\multirow{2}{*}{\textbf{Model}} & \multicolumn{2}{c}{\textbf{QuaRTz}} & \multicolumn{2}{c}{\textbf{QuaRel}} \\
\cmidrule(l){2-5} 
 & Dev         & Test         & Dev          & Test  \\\midrule
Random & 50.0 & 50.0 &50.0 &50.0 \\
QUASP  &- & - & 62.1 & 56.1\\
QUASP+  &- & - & 68.9 & 68.7\\\midrule
BERT & - & 67.7 & - & 53.0 \\
BERT(PFT on RACE) & - & 79.8 & - & 79.9\\
gvQPS+  &- & - & - & 76.6\\
DeepEKR   & - & 79.8 & - & 81.15\\\midrule
RoBERTa & 86.8$^{*}$ & 85.7$^{*}$& 81.1(84.5$^{*}$) & 80.0(83.3$^{*}$) \\
LG(DA) & - & - & 84.5 & 84.7 \\
LG(DA+Reg) & - & - & 85.1 & 85.0 \\
Ours & \textbf{89.6} & \textbf{89.9} & \textbf{89.5} & \textbf{86.8} \\
\bottomrule[2pt]
\end{tabular}%
}
\caption{Performance of different models on both datasets, where PFT stands for pre-finetune, and $*$ means the result we implemented.}
\label{tab:QA_performance}
\end{table}
\section{Experimental Results}
\subsection{Question Answering Performance}
The performance of the development and test of QuaRTz and QuaRel is shown in Table \ref{tab:QA_performance}. Our model outperformed the existing state-of-the-art models (RoBERTa and LG with DA+Reg) by a large margin (2.8\% and 4.2\% absolute difference on QuaRTz, 4.4\%, and 1.8\% on QuaRel). These results illustrate that compared to the semantic parser-based model and the one-step ``black box" model, our approach that imitates human cognitive behaviors is more capable of conducting qualitative reasoning and generalization.

Furthermore, compared with the semantic parser-based methods (i.e., the first group in the table), almost all language model-based methods achieved superior performance, demonstrating the power of the pre-trained language model. However, these models are mostly data-driven. The BERT model without pre-finetuning on RACE~\cite{lai-etal-2017-race} was unable to converge on QuaRel. Furthermore, the data augmentation technique resulted in a 4.7\% improvement in the LG(DA) model, proving the one-step ``black box" model has high demand for training resources. Nonetheless, our approach, without any additional training data, still achieves better performance, demonstrating its effectiveness and rationality.

\subsection{Reasoning Component Performance}
In the Reasoning component, we designed multiple neural network modules for two reasoning chains. Each module was designed to accomplish its sub-task and provide an intermediate prediction that explains the reasoning process in a human cognitive manner. To evaluate the effectiveness of each module, we converted the attention vectors over the context into a text span and the probability scores into a predicted class. The conversion details are in the Appendix~\ref{sec:CI}. Table~\ref{tab:module_performance} shows the performance results for each module.
\begin{table}[t]
\resizebox{\linewidth}{!}{%
\begin{tabular}{@{}lcccc@{}}
\toprule[2pt]
\textbf{Module}                        &         & \textbf{F1} & \textbf{Fuzzy F1} & \textbf{Accuracy} \\ \midrule
\multirow{2}{*}{Find Cause and Effect} & $\bm{p}_{c}^{b}$   & 72.6        & 82.3              & -                 \\
                                       & $\bm{p}_{e}^{b}$  & 67.0        & 78.4              & -                 \\
Polarity Check                         & $\bm{p}_{pol}$       & -           & -                 & 88.8              \\ \midrule
Find World                             & $\bm{p}_{w}^{s}$   & 76.4        & 82.5              & -                 \\ 
Value Prediction                       & $\bm{p}_{val}$       & -           & -                 & 91.5              \\\midrule
\multirow{2}{*}{Find Worlds}           & $\bm{p}_{w1}^{s}$ & 77.3        & 83.4              & -                 \\
                                       & $\bm{p}_{w2}^{s}$ & 74.9        & 80.6              & -                 \\
World Comparison                       & $\bm{p}_{com}$       & -           & -                 & 84.6              \\ \midrule
Reasoning Type Classification          & $\bm{p}_{cla}$       & -           & -                 & 88.0              \\ \bottomrule[2pt]
\end{tabular}%
}
\caption{Performance in the Reasoning component.}
\label{tab:module_performance}
\end{table}

\textit{Find Cause and Effect}, \textit{Find World}, and \textit{Find Worlds}, three modules that should detect concerned text spans from the context, all achieved adequate F1 scores. There are two reasons why our model did not score very well on a span extraction task like extractive machine reading comprehension (e.g., SQuAD\footnote{\url{https://rajpurkar.github.io/SQuAD-explorer/}}). First, the text span in our task is not as accurately defined as the answer in MRC. Generally, the length of the text span is long, and the boundary is fuzzy. Second, our approach computed the attention score for each token in the context and leveraged it softly, and thus less sensitive to boundary detection. However, when we used fuzzy F1 scores as the evaluation metrics (introduced in~\cite{ren-etal-2020-towards}, which were marked as 1 as long as the original F1 was not equal to 0), the scores for all modules increased by a large margin, proving the reasoning ability.

\textit{Polarity Check}, \textit{Value Prediction}, and \textit{World Comparison}, three modules that require classification capability based on the given span prediction of the upstream modules, achieved high accuracy (88.8\%, 91.5\%, and 84.6\%), indicating the rationality of our end-to-end sequential model design. Additionally, We could argue that determining whether an attribute is incremental or decremental is a relatively simple task for language models.

The \textit{Reasoning Type Classification} module obtained a good-enough performance (90.1\%), the recall value for Prediction type was 88.0\%, and the recall value for Comparison type was 76.7\%. The result is acceptable as some questions could be solved simultaneously in a Prediction and a Comparison manner. For example, the Comparison question in Figure 1 can also be considered a Prediction if we care only about the world with high priority (i.e., \textit{two planets with very large mass}). This phenomenon somehow increased the fault tolerance and improved the generality of our method.

\begin{figure}[t]
    \includegraphics[width= \linewidth]{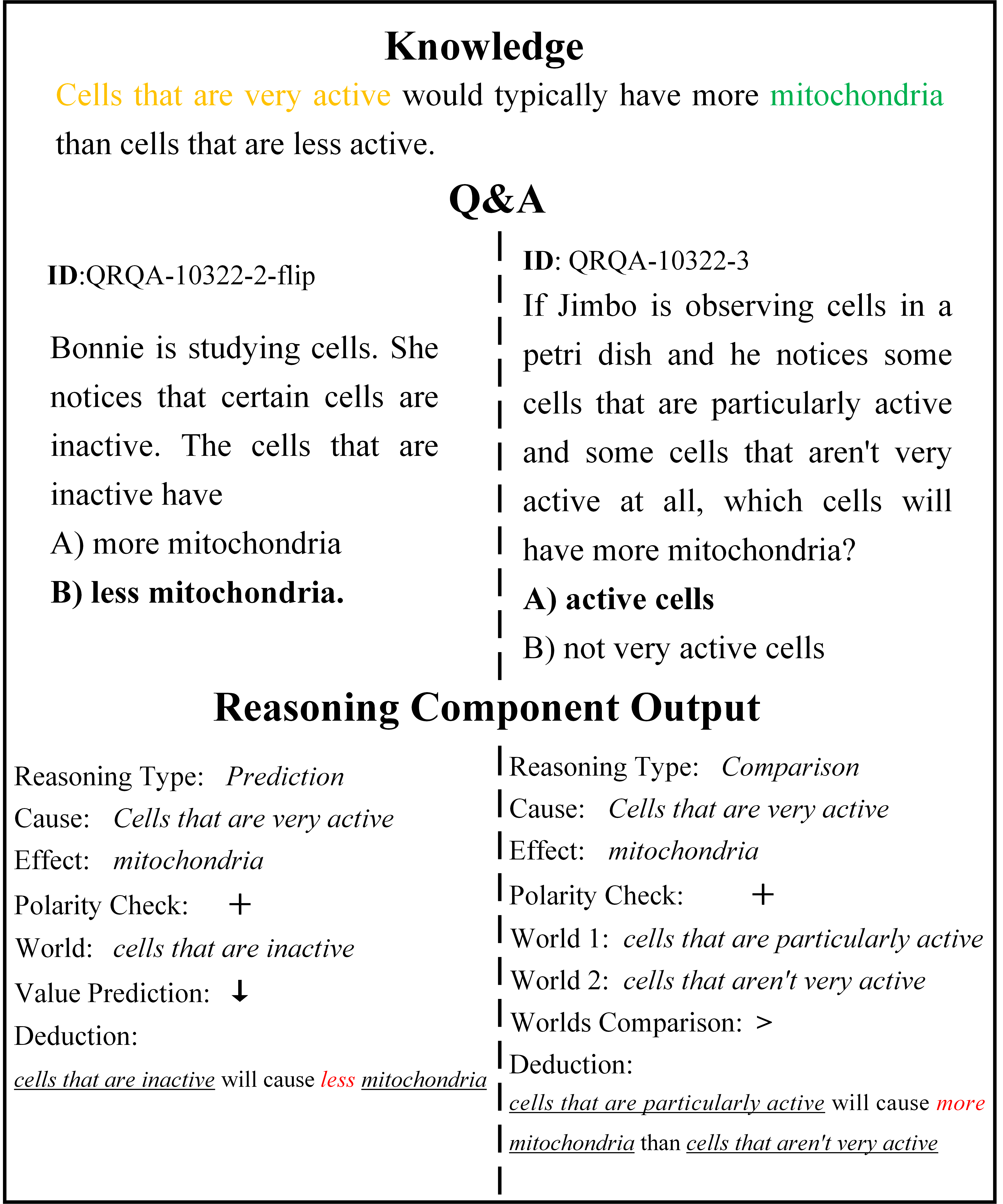}
    \caption{An example from QuaRTz with visualized intermediate outputs provided by our model.}
    \label{fig:case}
\end{figure}
\paragraph{Case study:}
Our approach provides intermediate outputs that sufficiently explain the reasoning process, which can lead to interpretability. Figure~\ref{fig:case} presents examples of the transparent reasoning process run by our model. More examples can be found in the Appendix~\ref{sec:more_ex}. The knowledge states the relationship between \textit{active cells} and \textit{mitochondria}, while two questions describe different scenarios to test the application of this knowledge.

As shown in Figure~\ref{fig:case}, our model outputs several intermediate results through the Reasoning component. First, it identifies the reasoning chain type for each question; the left one is prediction, while the right one requires comparison. Then, each module in the corresponding reasoning chain outputs its prediction. For example, the \textit{Find Cause and Effect} module captured the cause property and the effect property in the knowledge. Then the \textit{Polarity Check} module correctly figured out the positive relationship between them. The examples indicate that our approach not only conducts final answer prediction but also explains the machine understanding and reasoning process.

\subsection{Human Evaluation}
In the Answer Prediction component, we fed the generated synthetic text instead of given knowledge into the model. To further measure the effect of synthetic text on answer prediction, we introduced the human evaluation widely used in NLG field. We assembled ten well-educated volunteers and gave each person 30 randomly sampled questions from QuaRTz, half of which provided knowledge, half of which provided our synthetic text. Then we asked the participants to rate the question on a 5-point scale for each following metric:
\textit{fluency} (does it read coherently), \textit{informativeness} (how much information is contained), \textit{explicitness} (does it describe the relation clearly), and \textit{relevance} (is it relevant to the question). We also measured the time consumption for problem-solving by giving knowledge or synthetic text. Table~\ref{tab:human} shows the results.

The given knowledge had higher scores for \textit{fluency} and \textit{informativeness} because the knowledge sentences were manually extracted from a large corpus and may contain multiple relations among multiple properties, while the synthetic text was generated by slot-filling and described only the relation mentioned in the question. Furthermore, the synthetic text achieved superior scores for \textit{explicitness} and \textit{relevance}, which indicates our Reasoning component is successful in understanding and reasoning between the question and knowledge and expressed the explicit relationship in a way close to the description in the question scenario. Additionally, this is shown in that the participants took less time to answer questions with synthesized text.
\begin{table}[t]
\resizebox{\linewidth}{!}{%
\begin{tabular}{@{}lcc@{}}
\toprule
                             & \textbf{Given Knowledge} & \textbf{Synthetic Text} \\\midrule
\textbf{Fluency}             & \textbf{4.3}             & 3.9                     \\
\textbf{Informativeness}     & \textbf{4.5}             & 3.3                     \\
\textbf{Explicitness}        & 3.4                      & \textbf{4.4}            \\
\textbf{Relevance}           & 3.8                      & \textbf{4.6}            \\
\textbf{Time Consumption(s)} & 31.1                     & \textbf{24.3}    \\\bottomrule      
\end{tabular}%
}
\caption{Human evaluation results.}
\label{tab:human}
\end{table}
\subsection{Error Analysis}
We randomly sampled 60 examples for error analysis and more than half were caused by wrong predictions in the Reasoning component (48 out of 60). The other errors were in two main categories.
\paragraph{Incomplete Knowledge:}
These errors occurred when the question mentioned only a world without any attribute value description; e.g., the question aimed to predict the gravitational pull for the Sun, and knowledge told that a larger mass causes a larger gravitational pull. To answer this question, we need to know that the sun has a large mass.
\paragraph{Incomplete Synthetic Text:}
Two reasons cause such errors. One is that the boundaries are not well-defined when attention vectors are turned into text, resulting in a lack of fluency or loss of information. The other is a mismatch between the knowledge and the question; e.g., the knowledge talks about the distance, but the question describes hugging. Thus, the generated synthetic text does not provide enough information to solve the problem.

\section{Conclusion}
In this paper, we aimed to solve the qualitative reasoning task in an interpretable manner. Inspired by human cognition, we first summarized the questions into two categories, Prediction and Comparison. Then an end-to-end trained Reasoning component that contains two reasoning chains was designed. Both reasoning chains contained multiple neural modules that provide transparent intermediate predictions for the understanding and reasoning process. The experimental results showed the effectiveness of our approach, and the analysis of each module and case study demonstrated the superior interpretability compared with the ``black-box" model. Moreover, we found that some questions could be solved by both reasoning chains, thus increasing the default tolerance and generalization capability. Furthermore, a human evaluation was conducted to validate the function of the synthetic text and provide an additional explanation for the superior performance achieved by our method. However, the error analysis showed the inadequacy under complicated scenarios. Therefore, our future work will focus on applying interpretable reasoning on complex reasoning tasks. The annotated data and models are shared publicly~\footnote{\url{https://github.com/Borororo/Interpretable_QR}}.
\section*{Acknowledgement}
We acknowledge this work is supported by the Joint Funds of the National Natural Science Foundation of China (Grant No. U19B2020).
\bibliography{anthology,acl2020}
\bibliographystyle{acl_natbib}

\appendix

\section{Appendices}
\label{sec:appendix}

\subsection{Parameters List}
\label{sec:p_list}
Table~\ref{tab:params_R} and Table~\ref{tab:params_AP} show the parameter details used in our settings.
\begin{table}[h!]
\resizebox{\linewidth}{!}{%
\begin{tabular}{@{}lcc@{}}
\toprule
\textbf{Reasoning }  &Search Space(Bounds)  & Best Assignment \\ \midrule
Max Seq. Length            & \textit{choice}[256,384,512]                   & 512              \\
Learning Rate              & \textit{uniform-float}[5e-6,3e-5]                    & 1e-5          \\
Batch Size per GPU         & \textit{choice}[4,8,16]                       & 8                \\
Gradient Accumulation  & \textit{choice}[1,2]                       & 2                 \\
No. of Epoch               & \textit{uniform-integer}[1,10]                       & 8                 \\ \midrule
No. of Search trials        & 20                      & 20 \\
Optimizer    & Adam                   &Adam \\
Fixed Length for B,S     & [200,200]             & [200,200]                \\ 
No. of GPU(RTX8000)                 & 1                      & 1                \\
Average runtime (mins)     & 25                       & 25      \\
\bottomrule
\end{tabular}%
}
\caption{Detailed parameters used in the Reasoning component.}
\label{tab:params_R}
\end{table}

\begin{table}[h!]
\resizebox{\linewidth}{!}{%
\begin{tabular}{@{}lcc@{}}
\toprule
\textbf{Answer Prediction}         &Search Space(Bounds)  & Best Assignment \\ \midrule
Max Seq. Length            & \textit{choice}[256,384,512]                  & 256              \\
Learning Rate              & \textit{uniform-float}[1e-6,3e-5]                    & 1e-5            \\
Batch Size per GPU         & \textit{uniform-integer}[4,8,16]                       & 16                 \\
Gradient Accumulation  & \textit{choice}[1,2]                       & 2                 \\
No. of Epoch               & \textit{uniform-integer}[5,15]                        & 10                \\ \midrule
No. of Search trials        & 30                    &3 0 \\
learning rate optimizer    & Adam                   &Adam \\
No. of GPU(RTX8000)                & 2                      & 2                 \\
Average runtime (mins)     & 40                       &40      \\
\bottomrule
\end{tabular}%
}
\caption{Detailed parameters used in the Answer Prediction component.}
\label{tab:params_AP}
\end{table}

\subsection{Auxiliary Supervision Construction}
\label{sec:ASC}
\begin{figure}[t]
    \includegraphics[width= \linewidth]{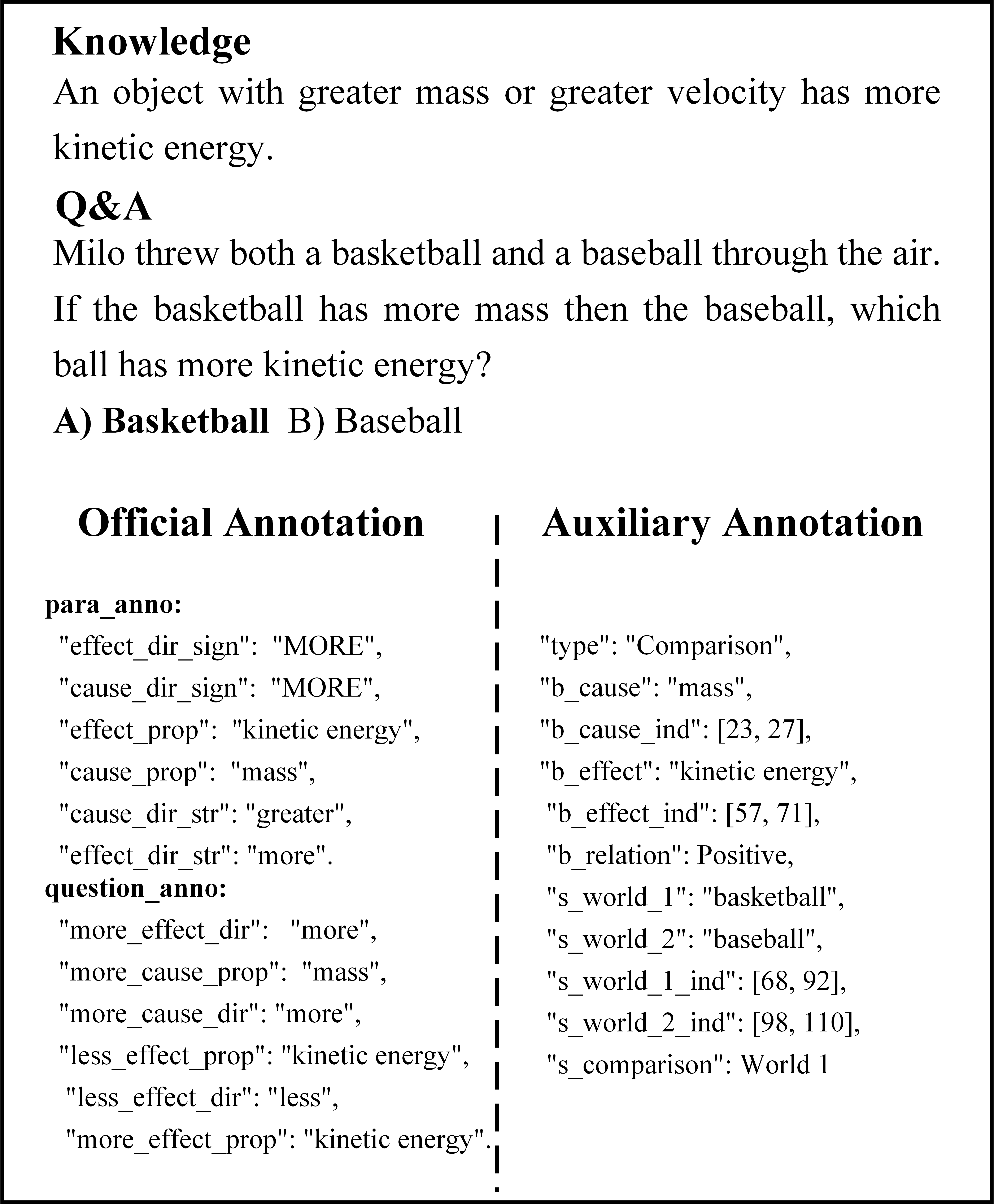}
    \caption{An example with auxiliary supervision.}
    \label{fig:label_example}
\end{figure}

Figure~\ref{fig:label_example} shows one labelled example. The official dataset already provides detailed annotations for both knowledge and question, where some annotations could be directly introduced in our training objective. For the example, ``para\_anno" tells the cause and effect properties described in the knowledge, and the polarity among them would be marked as positive if the ``effect\_dir\_sign" is the same as the ``casue\_dir\_sign", otherwise will be marked as negative.

However, some annotations are still missing, e.g., reasoning type. We introduce two methods to get additional labels. On the one hand, We made some hypotheses. For example, if ``question\_anno" contains ``more\_effect\_xxx" and ``less\_effect\_xxx" simultaneously, we would consider this question as a comparison type. On the other, we annotate the samples manually. For the questions without qualified annotation, we label them by hand. Particularly, we find the text spans in the knowledge and question, then transform the annotations to the machine-readble form automatically.

\subsection{Conversion Instruction}
\label{sec:CI}
For \textit{Find Cause and Effect}, \textit{Find World}, and \textit{Find Worlds} modules, we should convert the attention vector output into a text span. 

First, the token with the highest probability is selected. After that, we check left and right from the selected word position and extend it if the probability of each neighbor is larger than the threshold. The threshold value is determined based on F1 scores.

For the rest modules, we select the predicted class with the highest probability.

\subsection{Ablation Study on sub-tasks}
\label{sec:albation}
We decompose the complex qualitative reasoning task into several simple sub-tasks according to logic and then train several sub-tasks together. Naturally, as the number of tasks increases, errors accumulate, but our method ultimately performs better. To further examine the rationality of our work, we conduct an ablation study on partial modules. 

\begin{table}[h]
\resizebox{\linewidth}{!}{%
\begin{tabular}{@{}lccccc@{}}
\toprule
\multirow{2}{*}{\textbf{Modules}} & \multicolumn{2}{c}{\textbf{Cause}} & \multicolumn{2}{c}{\textbf{Effect}} & \textbf{Polarity} \\ \cmidrule(l){2-6} 
                                  & F1               & Fuzzy F1        & F1               & Fuzzy F1         & Accuracy          \\ \midrule
\textbf{Find Cause and Effect + Polarity Check}                  & \textbf{72.6}    & \textbf{82.3}   & \textbf{67.0}    & \textbf{78.4}    & \textbf{88.8}     \\
\quad \textbf{only Find Cause and Effect}                & 71.7             & 81.8            & 66.8             & 75.5             & 58.6              \\
\quad \textbf{only Polarity Check}            & 8.3              & 14.3            & 9.6              & 16.7             & 83.3              \\ \bottomrule
\end{tabular}%
}
\caption{Ablation study on Find Cause and Effect and Polarity Check modules.}
\label{tab:ablation}
\end{table}

Table~\ref{tab:ablation} shows the ablation study result on \textit{Find Cause and Effect} and \textit{Polarity Check} modules. The model achieves the best performance when we train two modules together. However, when we remove any module, i.e., focus on a single sub-task, the corresponding performance deteriorates. This may due to our approach, which mimics human cognitive design, enables the model to make better use of information from upstream modules to facilitate training. Therefore, our model could finally achieve superior performance.

\subsection{More Examples}
\label{sec:more_ex}
We present more example with intermediate outputs that correctly answered by our model in Table~\ref{tab:more_ex}.
\newpage

\begin{table*}[t]
\resizebox{0.95\textwidth}{!}{%
\begin{tabular}{p{0.6\textwidth}p{0.35\textwidth}}
\toprule[2pt]
ID:QRQA-10004-2 \newline
\textbf{Knowledge}: The gravitational force increases with mass and decreases with the distance between the bodies. \newline
\textbf{Q\&A}: John was watching the physics calculator and noted a profound finding. As the mass increases, the pull of the gravitational force A) Decreases \textbf{B) Increases}. & 
Type: \textit{Prediction} \newline
Cause:  \textit{mass} \newline
Effect: \textit{gravitational force} \newline
Polarity: \textbf{+}\newline
World: \textit{mass increases} \newline
Value:  $\uparrow $ \newline
Deduction: \newline
\textit{mass increases} will cause \textbf{more} \textit{gravitational force}.\\ \midrule
 ID:QRQA-10004-2-flip \newline
\textbf{Knowledge}: The gravitational force increases with mass and decreases with the distance between the bodies. \newline
\textbf{Q\&A}: John was watching the physics calculator and noted a profound finding. As the mass decreases, the pull of the gravitational force A) Decreases \textbf{B) Increases}.& 
Type: \textit{Prediction} \newline
Cause:  \textit{mass} \newline
Effect: \textit{gravitational force} \newline
Polarity: \textbf{+}\newline
World: \textit{mass decreases} \newline
Value:  $\downarrow $ \newline
Deduction: \newline
\textit{mass decreases} will cause \textbf{less} \textit{gravitational force}.\\ \midrule 
 ID:QRQA-10228-1 \newline
\textbf{Knowledge}: The larger the light collecting area, the more light a telescope gathers and the higher resolution (ability to see fine detail) it has.\newline
\textbf{Q\&A}: Compared to a 1 inch wide telescope, would a 100 meter telescope collect \textbf{A) more light} B) less light?& 
Type: \textit{Comparison} \newline
Cause:  \textit{collecting area} \newline
Effect: \textit{light} \newline
Polarity: \textbf{+}\newline
World 1: \textit{1 inch wide telescope} \newline
World 2: \textit{100 meter telescope} \newline
Comparison:  $<$ \newline
Deduction: \newline
\textit{1 inch wide telescope} will cause \textbf{less} \textit{light} than \textit{100 meter telescope}.\\ \midrule 
 ID:QRQA-10228-1-flip \newline
\textbf{Knowledge}: The larger the light collecting area, the more light a telescope gathers and the higher resolution (ability to see fine detail) it has.\newline
\textbf{Q\&A}: Compared to a 100 meter wide telescope, would a 1 inch telescope collect A) more light \textbf{B) less light}? & 
Type: \textit{Comparison} \newline
Cause:  \textit{collecting area} \newline
Effect: \textit{light} \newline
Polarity: \textbf{+}\newline
World 1: \textit{100 meter wide telescope} \newline
World 2: \textit{1 inch telescope} \newline
Comparison:  $>$ \newline
Deduction: \newline
\textit{100 meter wide telescope} will cause \textbf{more} \textit{light} than \textit{1 inch telescope}.\\ \bottomrule[2pt]
\end{tabular}%
}
\caption{}
\label{tab:more_ex}
\end{table*}

\end{document}